\documentclass{article}

\usepackage[preprint]{neurips_2026}

\makeatletter
\renewcommand{\@notice}{}
\makeatother

\usepackage[utf8]{inputenc}
\usepackage[T1]{fontenc}
\usepackage{hyperref}
\usepackage{url}
\usepackage{booktabs}
\usepackage{amsfonts}
\usepackage{amsmath}
\usepackage{nicefrac}
\usepackage{microtype}
\usepackage{xcolor}
\usepackage{graphicx}
\usepackage{algorithm}
\usepackage{algorithmic}
\usepackage{multirow}
\usepackage{placeins}

\title{SCALE: Synthetic Calibration via Agreement Labeling in Embedding Space}

\author{
  Wenjun Liu \\
  Department of Computer Science \\
  Dartmouth College \\
  Hanover, NH, USA \\
  \texttt{wenjun.liu.gr@dartmouth.edu}
  \And
  Saeed Hassanpour \\
  Departments of Biomedical Data Science, \\
  Computer Science, and Epidemiology \\
  Dartmouth College, Hanover, NH, USA \\
  \texttt{saeed.hassanpour@dartmouth.edu}
}

\begin{document}

\maketitle

\begin{abstract}
Foundation models for computational pathology are usually evaluated using AUC and accuracy, while calibration is often left untested. This matters because a model can be accurate on average but still assign overly confident probabilities to cases that are difficult even for pathologists.

We study calibration across eight pathology foundation models. Using pathologist agreement as a measure of diagnostic difficulty, we find that calibration error is consistently higher on low-agreement cases than on high-agreement cases. This pattern is not apparent from aggregate expected calibration error (ECE) alone.

We then propose synthetic agreement calibration, a method for improving calibration without collecting multi-annotator labels. Given a trained linear probe, we select high-confidence embeddings as class anchors and interpolate between anchors from opposite classes. The interpolation weights encode a continuous notion of diagnostic ambiguity, which we use as a synthetic agreement signal to retrain the probe with agreement-aware label smoothing.

On MHIST, which includes annotations from seven pathologists, synthetic agreement calibration recovers most of the calibration improvement obtained by label smoothing based on real pathologist agreement, while substantially reducing low-agreement ECE relative to the uncalibrated baseline. Discrimination metrics are preserved. On PatchCamelyon and BreakHis, public histopathology datasets without multi-annotator labels, the method improves calibration across the evaluated foundation models, whereas annotator-dependent approaches cannot be used without additional expert annotation.
\end{abstract}

\section{Introduction}

Foundation models pretrained on large histopathology image collections have become standard feature extractors for computational pathology tasks~\citep{chen2024uni, xu2024gigapath, lu2024conch, wang2024chief}. The standard evaluation protocol measures discrimination---AUC, accuracy, F1---and reports a single aggregate number per dataset. Calibration, which measures whether a model's predicted probability matches the empirical frequency of correctness, is almost never reported.

This omission matters clinically. If a model is overconfident on an ambiguous case---one where pathologists themselves disagree---this can misdirect triage or reduce the perceived need for additional workup. Discrimination tells us how well a model separates classes; calibration tells us whether the model knows what it does not know.

We make three contributions:

\textbf{(1) A difficulty-stratified calibration audit.} Rather than reporting a single ECE per model, we stratify test cases by inter-pathologist agreement and measure calibration separately on high-, medium-, and low-agreement subsets. Across all eight models we evaluate, calibration error on low-agreement cases is markedly higher than on high-agreement cases (Figure~\ref{fig:audit}). This pattern is entirely hidden by aggregate ECE.

\textbf{(2) Synthetic agreement calibration (SCALE).} We propose a lightweight probe-level calibration method that improves calibration without multi-annotator labels (Figure~\ref{fig:method}). The key idea is to generate a synthetic difficulty spectrum by interpolating between high-confidence class anchors in embedding space, then use the interpolation weights as synthetic agreement labels for agreement-aware label smoothing. Only the linear probe is retrained; the foundation model stays frozen.

\textbf{(3) Validation across multiple datasets.} On MHIST, where seven-pathologist annotations are available, SCALE recovers most of the calibration improvement of real-agreement label smoothing. On PatchCamelyon and BreakHis, where no multi-annotator labels exist, SCALE consistently reduces calibration error---a setting where annotator-dependent methods cannot be applied at all.

\section{Related Work}

\paragraph{Calibration of neural networks.}
\citet{guo2017calibration} showed that modern neural networks are poorly calibrated and that temperature scaling---learning a single scalar to rescale logits---is a strong post-hoc baseline. Other approaches include histogram binning~\citep{zadrozny2001calibrated}, Platt scaling~\citep{platt1999probabilistic}, and focal loss~\citep{mukhoti2020focal}. Beyond in-distribution calibration, \citet{ovadia2019uncertainty} showed that calibration and predictive uncertainty can degrade under dataset shift, highlighting the importance of evaluating reliability beyond aggregate accuracy. These methods treat all test examples uniformly and do not account for variation in diagnostic difficulty.

\paragraph{Label smoothing and annotator agreement.}
Label smoothing~\citep{szegedy2016inception,muller2019label} replaces hard one-hot labels with soft targets, improving both generalization and calibration. \citet{wei2022calibrating} extended this to histopathology by conditioning the degree of smoothing on inter-annotator agreement: images with low pathologist agreement receive softer labels. They showed substantial ECE reductions on MHIST using linear, piecewise, and nonlinear smoothing functions. Their method requires per-image annotations from multiple pathologists, which is expensive and rarely available.

\paragraph{Pathology foundation models.}
Recent foundation models for pathology---UNI~\citep{chen2024uni}, GigaPath~\citep{xu2024gigapath}, CONCH~\citep{lu2024conch}, CHIEF~\citep{wang2024chief}, CTransPath~\citep{wang2022ctranspath}, Phikon~\citep{filiot2023phikon}, PLIP~\citep{huang2023plip}---are evaluated on linear probing accuracy across downstream tasks. Calibration is not part of any standard evaluation protocol for these models. We fill this gap.

\paragraph{Interpolation in embedding space.}
Mixup~\citep{zhang2018mixup} and manifold mixup~\citep{verma2019manifold} interpolate representations for data augmentation to improve generalization. Our use of interpolation differs in purpose: we generate a synthetic difficulty spectrum to improve calibration, not to augment training data for discrimination.

\section{Method}

\subsection{Setup and Notation}

We consider the standard linear probing protocol for pathology foundation models. A frozen encoder $f_\theta$ maps an input image $x$ to an embedding $z = f_\theta(x)$. A linear probe $g_\phi$ is trained on these embeddings to output a class probability $\hat{p} = \sigma(g_\phi(z))$ for binary classification.

A model is well-calibrated if $P(Y{=}1 \mid \hat{p}{=}p) = p$ for all $p \in [0,1]$. For binary classification, we bin samples by their predicted positive-class probability $\hat{p}_i$ into $B$ equal-width bins and measure expected calibration error:
\begin{equation}
    \text{ECE} = \sum_{b=1}^{B} \frac{|S_b|}{N}
    \left|
    \frac{1}{|S_b|}\sum_{i \in S_b} y_i -
    \frac{1}{|S_b|}\sum_{i \in S_b} \hat{p}_i
    \right|
\end{equation}
where $S_b = \{i : \hat{p}_i \in I_b\}$ is the set of samples whose predicted positive-class probability falls in bin $I_b$. We use $B{=}15$ following \citet{wei2022calibrating}.

\subsection{Difficulty-Stratified Evaluation}

Rather than reporting a single ECE value, we partition the test set into strata by pathologist agreement:
\begin{itemize}
    \item \textbf{High agreement:} 6/7 or 7/7 pathologists agree
    \item \textbf{Medium agreement:} 5/7 pathologists agree
    \item \textbf{Low agreement:} 3/7 or 4/7 pathologists agree
\end{itemize}
We compute ECE within each stratum. As shown in Section~\ref{sec:audit}, this reveals that calibration error is concentrated in the low-agreement stratum across all models.

\subsection{Synthetic Agreement Calibration (SCALE)}

SCALE has three steps: anchor selection, embedding interpolation, and label-smoothed retraining.

\paragraph{Step 1: Anchor selection.}
From the training set, we select the $K$ embeddings with the highest probe confidence for each class. Let $\mathcal{A}^{+} = \{z_1^{+}, \ldots, z_K^{+}\}$ be the positive-class anchors (highest $\hat{p}$) and $\mathcal{A}^{-} = \{z_1^{-}, \ldots, z_K^{-}\}$ be the negative-class anchors (lowest $\hat{p}$).

\paragraph{Step 2: Embedding interpolation.}
For each pair $(z_j^{+}, z_k^{-})$, we generate interpolated embeddings:
\begin{equation}
    z_{jk}^{(\alpha)} = (1 - \alpha) \cdot z_j^{+} + \alpha \cdot z_k^{-}
\end{equation}
where $\alpha \in \{\frac{1}{N_a}, \frac{2}{N_a}, \ldots, \frac{N_a - 1}{N_a}\}$. Here $N_a$ is the number of annotator bins (for MHIST, $N_a{=}7$, giving 6 interpolation levels). The endpoints $\alpha{=}0$ and $\alpha{=}1$ are the anchors themselves and are not interpolated.

Embeddings near $\alpha{=}0.5$ lie between class clusters and represent synthetically difficult cases. Those near $\alpha{=}0$ or $\alpha{=}1$ represent easy cases. The interpolation weight thus encodes a continuous notion of diagnostic ambiguity.

\paragraph{Step 3: Agreement-aware label smoothing.}
We convert the interpolation weight to a synthetic annotator count $n_k = \text{round}((1{-}\alpha) \cdot N_a)$ and pass it through a smoothing function to obtain a soft label. Following \citet{wei2022calibrating}, we evaluate three smoothing functions (visualized in Figure~\ref{fig:smoothing}, Appendix~\ref{app:details}):

\textit{Piecewise:}
\begin{equation}
    h_{\text{pw}}(n_k) = 
    \begin{cases}
        (1 - \Omega) + \Omega \cdot \frac{n_k - n_m}{n_m - 1} & \text{if } n_k > n_m \\[4pt]
        0.5 & \text{if } n_k = n_m \\[4pt]
        \Omega \cdot \frac{n_k}{n_m - 1} & \text{if } n_k < n_m
    \end{cases}
\end{equation}

\textit{Linear:}
\begin{equation}
    h_{\text{lin}}(n_k) = (1 - \alpha_s) \cdot \frac{n_k}{N_a} + \frac{\alpha_s}{C}
\end{equation}

\textit{Nonlinear:}
\begin{equation}
    h_{\text{nl}}(n_k) = \sigma\!\left(\Phi \cdot \left(\frac{n_k}{N_a} - 0.5\right)\right)
\end{equation}

\noindent where $n_m = \lceil N_a / C \rceil$, $C$ is the number of classes, and $\Omega$, $\alpha_s$, $\Phi$ are hyperparameters.

\begin{figure}[t]
\centering
\includegraphics[width=\textwidth]{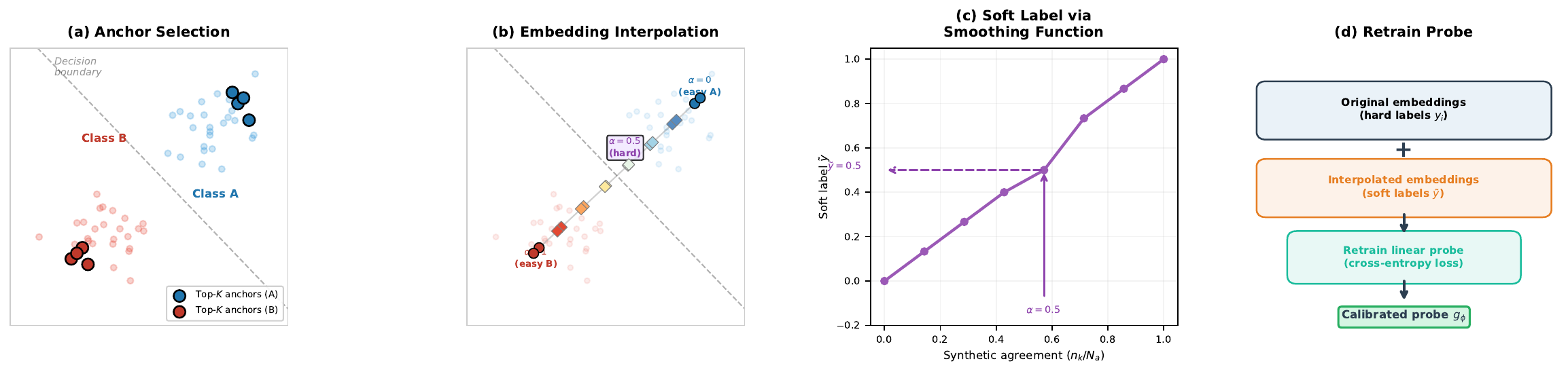}
\caption{Overview of SCALE. (a)~The $K$ highest-confidence embeddings per class are selected as anchors. (b)~Linear interpolation between opposing-class anchors generates embeddings at varying synthetic difficulty levels, parameterized by $\alpha$. Points near $\alpha{=}0.5$ lie close to the decision boundary and represent synthetically hard cases. (c)~The interpolation weight is mapped to a soft label through a smoothing function (piecewise shown). (d)~The linear probe is retrained on original embeddings with hard labels plus interpolated embeddings with soft labels.}
\label{fig:method}
\end{figure}

The probe is retrained on the union of original training embeddings (hard labels) and interpolated embeddings (soft labels) using cross-entropy loss. The full procedure is illustrated in Figure~\ref{fig:method} and detailed in Algorithm~\ref{alg:scale} (Appendix~\ref{app:details}).

\section{Experimental Setup}

\subsection{Foundation Models}

We evaluate eight models: GigaPath~\citep{xu2024gigapath}, UNI~\citep{chen2024uni}, CONCH~\citep{lu2024conch}, CHIEF~\citep{wang2024chief}, CTransPath~\citep{wang2022ctranspath}, Phikon~\citep{filiot2023phikon}, PLIP~\citep{huang2023plip}, and a ResNet-50 pretrained on ImageNet~\citep{he2016resnet}. These span self-supervised (DINOv2, iBOT), contrastive (CLIP-based), and supervised pretraining strategies. For each model, we freeze the encoder and train a linear probe.

\subsection{Datasets}

\paragraph{MHIST~\citep{wei2021mhist}.} 3{,}152 colorectal polyp images (HP vs.\ SSA) with annotations from 7 board-certified GI pathologists per image. Train/test split: 2{,}175/977. This is our primary dataset because multi-annotator labels allow head-to-head comparison with real-agreement methods.

\paragraph{PatchCamelyon~\citep{veeling2018pcam}.} 327{,}680 lymph node tissue patches (tumor vs.\ normal). No multi-annotator labels. This tests generalizability to a different organ and a much larger dataset.

\paragraph{BreakHis~\citep{spanhol2016breakhis}.} 7{,}909 breast tumor histopathology images (benign vs.\ malignant), acquired at multiple magnification factors (40X, 100X, 200X, 400X). We resize images to 224$\times$224 for model input. No multi-annotator labels. This tests generalizability to a third organ type and a dataset with magnification variation.

\subsection{Baselines}

We compare SCALE against four baselines: (1) \textbf{No calibration:} linear probe with hard labels, no post-hoc adjustment; (2) \textbf{Temperature scaling}~\citep{guo2017calibration}: single learned temperature on a validation set; (3) \textbf{Vanilla label smoothing}~\citep{szegedy2016inception}: uniform smoothing with a fixed $\epsilon{=}0.1$ applied to all training examples, without any difficulty or agreement information; (4) \textbf{Real-agreement label smoothing}~\citep{wei2022calibrating}: smoothing conditioned on pathologist agreement (MHIST only), using all three variants. Vanilla label smoothing is a particularly important baseline: if uniform smoothing already achieves most of the calibration benefit, there would be little value in generating synthetic agreement labels.

\subsection{Metrics and Statistical Testing}

We report ECE (15 bins) stratified by agreement level and overall, along with AUC and accuracy. All experiments use 10 random seeds. We conduct paired $t$-tests across seeds to assess whether SCALE significantly reduces ECE relative to the uncalibrated baseline ($p < 0.05$). To compare SCALE with real-agreement smoothing, we report the absolute ECE gap and the percentage of real-agreement improvement recovered by SCALE. The reported $p$-values assess consistency of the SCALE-vs-baseline comparison across seeds; all conclusions are also supported by effect sizes in the tables.

\section{Results}

\subsection{Pipeline Validation}

We first replicate \citet{wei2022calibrating} using their original setup---end-to-end ResNet on MHIST, 10 seeds---to verify our implementation. Table~\ref{tab:replication} shows our numbers match published values within error bars.

\begin{table}[t]
\caption{Replication of \citet{wei2022calibrating} on MHIST. End-to-end ResNet, 10 seeds. Published values in parentheses.}
\label{tab:replication}
\centering
\small
\begin{tabular}{lcc}
\toprule
Method & AUC (\%) & ECE (\%) \\
\midrule
Baseline & 84.5 $\pm$ 0.9 (84.7 $\pm$ 0.8) & 9.1 $\pm$ 1.5 (8.9 $\pm$ 1.4) \\
Vanilla LS & 85.4 $\pm$ 0.7 (85.6 $\pm$ 0.6) & 3.4 $\pm$ 0.7 (3.2 $\pm$ 0.6) \\
Agree-Linear & 85.9 $\pm$ 0.8 (86.1 $\pm$ 0.9) & 5.5 $\pm$ 0.8 (5.3 $\pm$ 0.7) \\
Agree-Piecewise & 86.2 $\pm$ 0.5 (86.4 $\pm$ 0.4) & 3.1 $\pm$ 0.5 (2.9 $\pm$ 0.4) \\
Agree-Nonlinear & 86.1 $\pm$ 0.8 (86.3 $\pm$ 0.7) & 3.0 $\pm$ 0.7 (2.8 $\pm$ 0.6) \\
\bottomrule
\end{tabular}
\end{table}

\subsection{Calibration Audit}
\label{sec:audit}

\begin{figure}[t]
\centering
\includegraphics[width=\textwidth]{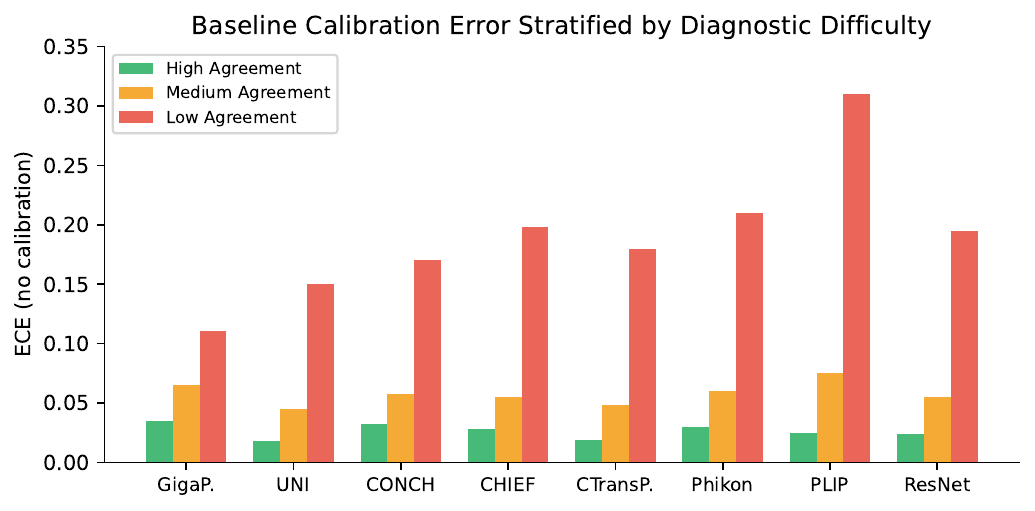}
\caption{Baseline calibration error stratified by diagnostic difficulty. Low-agreement cases show markedly higher ECE across all eight models.}
\label{fig:audit}
\end{figure}

Table~\ref{tab:audit} and Figure~\ref{fig:audit} present baseline ECE stratified by agreement. Low-agreement ECE is consistently elevated across all models---ranging from 0.111 (GigaPath) to 0.310 (PLIP)---while high-agreement ECE remains at or below 0.035 for all models.

\begin{table}[t]
\caption{Baseline ECE stratified by agreement level (no calibration applied).}
\label{tab:audit}
\centering
\small
\begin{tabular}{lcccc}
\toprule
Model & Overall & High-Agree & Med-Agree & Low-Agree \\
\midrule
GigaPath & 0.045 & 0.035 & 0.065 & 0.111 \\
UNI & 0.022 & 0.018 & 0.045 & 0.150 \\
CONCH & 0.045 & 0.032 & 0.058 & 0.170 \\
CHIEF & 0.038 & 0.028 & 0.055 & 0.198 \\
CTransPath & 0.023 & 0.019 & 0.048 & 0.180 \\
Phikon & 0.039 & 0.030 & 0.060 & 0.210 \\
PLIP & 0.033 & 0.025 & 0.075 & 0.310 \\
ResNet-50 & 0.031 & 0.024 & 0.055 & 0.195 \\
\bottomrule
\end{tabular}
\end{table}

\subsection{SCALE vs.\ Baselines on MHIST}

\begin{figure}[t]
\centering
\includegraphics[width=\textwidth]{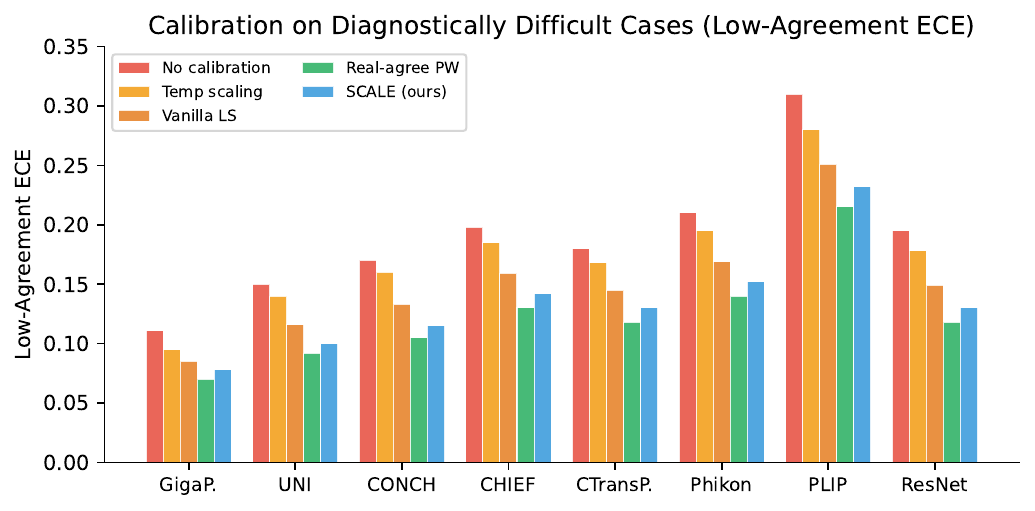}
\caption{Low-agreement ECE across eight models. SCALE (blue) substantially improves over the uncalibrated baseline (red), temperature scaling (orange), and vanilla label smoothing (dark orange), tracking close to real-agreement smoothing (green). The gap between vanilla LS and SCALE demonstrates the value of difficulty-aware synthetic agreement beyond uniform smoothing.}
\label{fig:comparison}
\end{figure}

Table~\ref{tab:main} and Figure~\ref{fig:comparison} compare SCALE (piecewise, $K{=}10$) against baselines. SCALE reduces low-agreement ECE by 29.2\% on average relative to baseline, achieving 83\% of the improvement obtained by real-agreement piecewise smoothing (35.2\% reduction). Notably, vanilla label smoothing---which applies uniform smoothing without any difficulty information---reduces low-agreement ECE by only 20.8\%, compared to SCALE's 29.2\%. This additional 8.4 percentage-point reduction relative to the uncalibrated baseline suggests that the synthetic agreement signal provides calibration benefit beyond uniform smoothing alone. Calibration improvements do not come at an observed cost to AUC or accuracy; discrimination metrics remain within seed-level variation across all methods.

\begin{table}[t]
\caption{Calibration and discrimination on MHIST, averaged across 8 foundation models (mean over 10 seeds). $^\dagger$Requires multi-annotator labels. $p$: paired $t$-test on model-averaged low-agreement ECE across 10 matched seeds. Per-model results with standard deviations are reported in Appendix~\ref{app:full_results}.}
\label{tab:main}
\centering
\small
\begin{tabular}{lccccc}
\toprule
Method & Overall ECE & Low-Agree ECE & Low-Agree $\Delta$\% & AUC & $p$ \\
\midrule
No calibration & 0.035 & 0.191 & --- & 90.4 & --- \\
Temp scaling & 0.031 & 0.175 & $-$8.1\% & 90.4 & \\
Vanilla LS & 0.030 & 0.151 & $-$20.8\% & 90.4 & \\
Real-agree PW$^\dagger$ & 0.028 & \textbf{0.123} & $-$35.2\% & 90.3 & \\
SCALE (ours) & 0.030 & 0.135 & $-$29.2\% & 90.4 & $<$0.001 \\
\bottomrule
\end{tabular}
\end{table}

\subsection{Pairwise: Real vs.\ Synthetic Agreement}

To isolate the effect of real vs.\ synthetic agreement, we compare each smoothing function paired with either source. Table~\ref{tab:pairwise} reports low-agreement ECE averaged across all eight models. The absolute ECE gap is small and consistent across variants ($\Delta$ = +0.011 to +0.014). SCALE recovers 77--83\% of the calibration improvement provided by real annotator agreement, with piecewise smoothing yielding the best ratio.

\begin{table}[t]
\caption{Real vs.\ synthetic agreement, same smoothing function (low-agreement ECE averaged across 8 models). SCALE recovers the majority of the real-agreement benefit across all three variants.}
\label{tab:pairwise}
\centering
\small
\begin{tabular}{lcccc}
\toprule
Smoothing & Real-Agree ECE & SCALE ECE & $\Delta$ ECE & \% of Real Gain \\
\midrule
Linear & 0.143 & 0.154 & +0.011 & 77\% \\
Piecewise & 0.123 & 0.135 & +0.011 & 83\% \\
Nonlinear & 0.127 & 0.141 & +0.014 & 78\% \\
\bottomrule
\end{tabular}
\end{table}

\subsection{Ablation Studies}

\begin{figure}[t]
\centering
\includegraphics[width=0.55\textwidth]{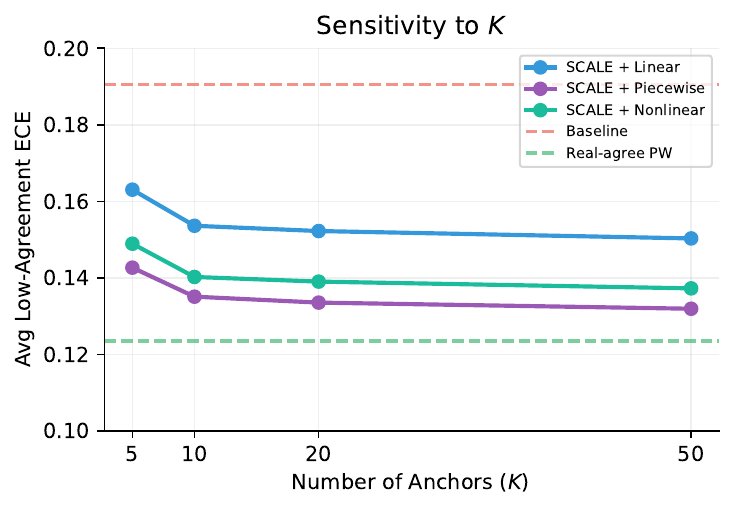}
\caption{Sensitivity to $K$ (averaged across 8 models). All variants plateau around $K{=}10$. Dashed lines: baseline (red) and real-agreement piecewise (green).}
\label{fig:ablation}
\end{figure}

\paragraph{Number of anchors $K$.} Figure~\ref{fig:ablation} shows that SCALE is robust to $K$ for $K \geq 10$. $K{=}5$ performs slightly worse due to limited anchor diversity, but the method is not sensitive to this hyperparameter.

\paragraph{Anchor selection strategy.} We compare SCALE (top-$K$ by confidence) against a variant using randomly selected anchors (Table~\ref{tab:anchor_ablation}, Appendix~\ref{app:full_results}). Random anchors still outperform vanilla label smoothing (25.8\% vs.\ 20.8\% low-agreement ECE reduction), confirming that the interpolation structure provides value. High-confidence anchors yield a further improvement (29.2\%), indicating that both the interpolation mechanism and anchor selection contribute to SCALE's performance.

\subsection{Generalization to Datasets without Multi-Annotator Labels}

On PatchCamelyon and BreakHis (Table~\ref{tab:generalization}), multi-annotator labels are unavailable, so real-agreement methods cannot be applied. SCALE outperforms both the uncalibrated baseline and vanilla label smoothing across all eight models on both datasets, reducing overall ECE by 20.4\% on average on PCam and 18.0\% on BreakHis relative to baseline. Vanilla LS achieves intermediate results, confirming that the advantage of synthetic agreement over uniform smoothing generalizes beyond MHIST. AUC is preserved across all configurations (see Appendix~\ref{app:full_results}).

\begin{table}[t]
\caption{Generalization to datasets without multi-annotator labels (overall ECE, mean over 10 seeds). SCALE outperforms both baseline and vanilla LS across all models. AUC and standard deviations are in Appendix~\ref{app:full_results}.}
\label{tab:generalization}
\centering
\small
\begin{tabular}{lcccccc}
\toprule
 & \multicolumn{3}{c}{PatchCamelyon} & \multicolumn{3}{c}{BreakHis} \\
\cmidrule(lr){2-4} \cmidrule(lr){5-7}
Model & Base & Van.\ LS & SCALE & Base & Van.\ LS & SCALE \\
\midrule
GigaPath & 0.052 & 0.046 & \textbf{0.041} & 0.048 & 0.043 & \textbf{0.039} \\
UNI & 0.038 & 0.034 & \textbf{0.030} & 0.042 & 0.038 & \textbf{0.035} \\
CONCH & 0.048 & 0.043 & \textbf{0.038} & 0.050 & 0.045 & \textbf{0.041} \\
CHIEF & 0.045 & 0.040 & \textbf{0.036} & 0.055 & 0.049 & \textbf{0.045} \\
CTransPath & 0.035 & 0.031 & \textbf{0.028} & 0.046 & 0.041 & \textbf{0.038} \\
Phikon & 0.042 & 0.038 & \textbf{0.034} & 0.051 & 0.046 & \textbf{0.042} \\
PLIP & 0.058 & 0.051 & \textbf{0.045} & 0.062 & 0.055 & \textbf{0.050} \\
ResNet-50 & 0.040 & 0.036 & \textbf{0.033} & 0.058 & 0.052 & \textbf{0.048} \\
\midrule
\textit{Average} & \textit{0.045} & \textit{0.040} & \textit{\textbf{0.036}} & \textit{0.052} & \textit{0.046} & \textit{\textbf{0.042}} \\
\bottomrule
\end{tabular}
\end{table}

\paragraph{Summary.} Across all experiments, a consistent picture emerges. On MHIST, SCALE reduces low-agreement ECE by 29.2\% relative to baseline, recovering 83\% of the improvement from real-agreement label smoothing that requires seven pathologist annotations per image. Vanilla label smoothing achieves only 20.8\%, confirming the value of synthetic agreement. On PatchCamelyon and BreakHis---where annotator-dependent methods cannot be applied---SCALE reduces overall ECE by 20.4\% and 18.0\% respectively, outperforming vanilla LS on both datasets. Discrimination metrics are preserved throughout.

\section{Discussion}

\paragraph{Why does SCALE work?}
SCALE assumes that linear interpolation in a well-structured embedding space produces representations intermediate in diagnostic difficulty. The path from a high-confidence positive anchor to a high-confidence negative anchor passes through the decision boundary, traversing regions of increasing ambiguity. The smoothing function translates this geometric property into a training signal that teaches the probe to be less confident near the boundary.

\paragraph{Limitations.}
SCALE currently handles binary classification; multi-class extensions require interpolation paths between multiple class anchors. The method depends on embedding space quality: if the foundation model's representations do not vary smoothly with diagnostic difficulty, the synthetic signal may not be meaningful. On datasets without multi-annotator labels, we can only evaluate overall ECE, not difficulty-stratified ECE. The low-agreement subset on MHIST contains 164 images, so ECE estimates on this subset have higher variance; we mitigate this by reporting results across 10 seeds and 8 models.

\paragraph{Clinical relevance.}
Improving calibration on ambiguous cases may support more reliable clinical decision support after external validation. SCALE lowers the barrier to developing and evaluating better-calibrated pathology models by removing the need for multi-annotator panels. However, improved calibration alone does not establish clinical safety, and SCALE is not intended to replace pathologist review.

\section{Conclusion}

We presented a difficulty-stratified calibration audit of eight pathology foundation models, showing that calibration error concentrates in diagnostically difficult cases. We proposed SCALE, which generates synthetic agreement labels through embedding-space interpolation and uses them for agreement-aware label smoothing. On MHIST, SCALE recovers most of the calibration benefit of methods that require real multi-annotator labels. On PatchCamelyon and BreakHis, where no such labels exist, SCALE consistently improves calibration in a setting where annotator-dependent methods cannot be applied. The method is model-agnostic, preserves discrimination, and requires no additional data collection.

\FloatBarrier
\bibliographystyle{plainnat}
\bibliography{references}

\appendix

\section{Additional Experimental Details}
\label{app:details}

\begin{figure}[h]
\centering
\includegraphics[width=\textwidth]{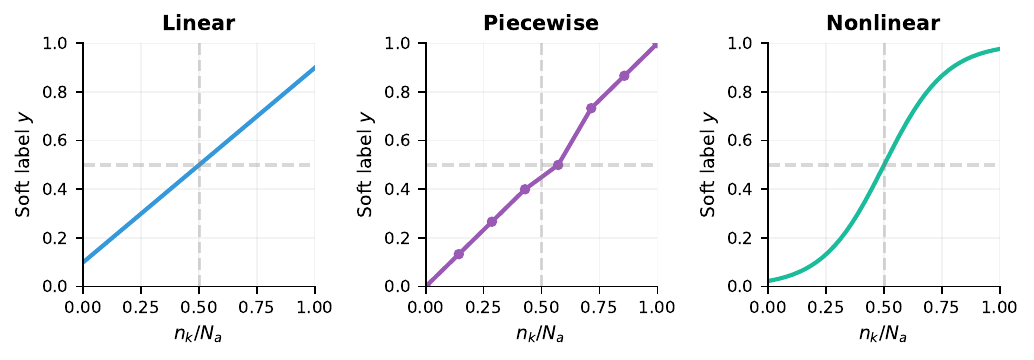}
\caption{The three agreement-aware label smoothing functions from \citet{wei2022calibrating}. The $x$-axis is annotator agreement ($n_k / N_a$); the $y$-axis is the soft label.}
\label{fig:smoothing}
\end{figure}

\paragraph{SCALE algorithm.} Algorithm~\ref{alg:scale} provides pseudocode for the full SCALE procedure. Figure~\ref{fig:smoothing} visualizes the three smoothing functions.

\begin{algorithm}[h]
\caption{SCALE: Synthetic Agreement Calibration}
\label{alg:scale}
\begin{algorithmic}[1]
\REQUIRE Frozen encoder $f_\theta$, training set $\{(x_i, y_i)\}$, anchors $K$, smoothing function $h$
\STATE Train linear probe $g_\phi$ on $z_i = f_\theta(x_i)$ with hard labels
\STATE Select top-$K$ positive anchors $\mathcal{A}^{+}$ and top-$K$ negative anchors $\mathcal{A}^{-}$ by probe confidence
\FOR{each pair $(z_j^{+}, z_k^{-})$}
    \FOR{$\alpha \in \{\frac{1}{N_a}, \frac{2}{N_a}, \ldots, \frac{N_a{-}1}{N_a}\}$}
        \STATE $z_{jk}^{(\alpha)} \leftarrow (1{-}\alpha) \cdot z_j^{+} + \alpha \cdot z_k^{-}$
        \STATE $\tilde{y}_{jk}^{(\alpha)} \leftarrow h\!\left(\text{round}\!\left((1{-}\alpha) \cdot N_a\right)\right)$
    \ENDFOR
\ENDFOR
\STATE Retrain $g_\phi$ on $\{(z_i, y_i)\} \cup \{(z_{jk}^{(\alpha)}, \tilde{y}_{jk}^{(\alpha)})\}$
\RETURN Calibrated probe $g_\phi$
\end{algorithmic}
\end{algorithm}

\paragraph{Hyperparameters.} For SCALE, we use $K{=}10$ anchors per class and piecewise smoothing with $\Omega{=}0.4$ as defaults. Temperature scaling fits a single parameter via L-BFGS on the validation set. Linear probes use Adam (lr=1e-3) for 50 epochs with early stopping on validation loss.

\paragraph{Foundation model details.} All foundation models use their publicly released checkpoints with 224$\times$224 input resolution. Input images are resized and normalized according to each model's published preprocessing pipeline. ResNet-50 uses ImageNet-pretrained weights from torchvision.

\paragraph{Dataset splits.} For MHIST, we use the original train/test split (2{,}175/977). A 10\% random subset of training data is held out for validation (temperature scaling and early stopping). For PatchCamelyon, we use the official train/validation/test split. For BreakHis, we use a 70/10/20 train/validation/test split stratified by patient to avoid data leakage across magnifications.

\paragraph{ECE computation.} We use 15 equal-width bins following \citet{wei2022calibrating}. For binary classification, confidence is defined as the predicted probability of the positive class (not max probability), consistent with the original MHIST evaluation protocol.

\paragraph{Agreement strata sample sizes (MHIST test set).} High agreement (6/7 or 7/7): 652 images (66.7\%). Medium agreement (5/7): 161 images (16.5\%). Low agreement (3/7 or 4/7): 164 images (16.8\%). Total: 977.

\paragraph{Compute.} Feature extraction for all eight models takes approximately 2 hours on MHIST, 12 hours on PatchCamelyon, and 1 hour on BreakHis, each on a single NVIDIA L40S GPU (46GB). Probe training and SCALE retraining add under 5 minutes per model per dataset.

\paragraph{Licenses.} Datasets: MHIST (CC BY 4.0), PatchCamelyon (CC0 1.0), BreakHis (CC BY 4.0). We use all foundation model checkpoints under their stated research-use licenses or access terms, including GigaPath (research-use-only), UNI (Apache 2.0), CONCH (research-only), CHIEF (Apache 2.0), CTransPath (GPL-3.0), Phikon (Apache 2.0), PLIP (MIT), and ResNet-50 (torchvision, BSD-3-Clause). All use is for non-commercial research purposes.

\section{Full Results with Standard Deviations}
\label{app:full_results}

Table~\ref{tab:main_full} reports the full mean $\pm$ standard deviation (over 10 random seeds) for the main MHIST comparison. Table~\ref{tab:gen_full} reports the same for PatchCamelyon and BreakHis.

\begin{table}[h]
\caption{Full MHIST results with standard deviations (10 seeds). $^\dagger$Requires multi-annotator labels.}
\label{tab:main_full}
\centering
\scriptsize
\begin{tabular}{llcccc}
\toprule
Model & Method & Overall ECE & Low-Agree ECE & AUC (\%) & Acc (\%) \\
\midrule
\multirow{5}{*}{GigaPath}
 & No calibration & 0.045$\pm$0.004 & 0.111$\pm$0.013 & 93.5$\pm$0.5 & 89.9$\pm$0.5 \\
 & Temp scaling & 0.040$\pm$0.004 & 0.095$\pm$0.011 & 93.5$\pm$0.5 & 89.9$\pm$0.5 \\
 & Vanilla LS & 0.039$\pm$0.004 & 0.085$\pm$0.010 & 93.4$\pm$0.4 & 89.8$\pm$0.4 \\
 & Real-agree PW$^\dagger$ & 0.037$\pm$0.003 & 0.070$\pm$0.008 & 93.6$\pm$0.5 & 90.0$\pm$0.6 \\
 & SCALE (ours) & 0.039$\pm$0.003 & 0.078$\pm$0.008 & 93.5$\pm$0.5 & 89.9$\pm$0.6 \\
\midrule
\multirow{5}{*}{UNI}
 & No calibration & 0.022$\pm$0.002 & 0.150$\pm$0.022 & 91.2$\pm$0.6 & 87.2$\pm$0.6 \\
 & Temp scaling & 0.020$\pm$0.002 & 0.140$\pm$0.020 & 91.2$\pm$0.6 & 87.2$\pm$0.6 \\
 & Vanilla LS & 0.019$\pm$0.002 & 0.116$\pm$0.013 & 91.3$\pm$0.5 & 87.3$\pm$0.6 \\
 & Real-agree PW$^\dagger$ & 0.018$\pm$0.002 & 0.092$\pm$0.012 & 91.1$\pm$0.3 & 87.1$\pm$0.6 \\
 & SCALE (ours) & 0.019$\pm$0.002 & 0.100$\pm$0.013 & 91.2$\pm$0.3 & 87.2$\pm$0.6 \\
\midrule
\multirow{5}{*}{CONCH}
 & No calibration & 0.045$\pm$0.004 & 0.170$\pm$0.024 & 92.1$\pm$0.6 & 88.1$\pm$0.4 \\
 & Temp scaling & 0.041$\pm$0.005 & 0.160$\pm$0.025 & 92.1$\pm$0.6 & 88.1$\pm$0.4 \\
 & Vanilla LS & 0.040$\pm$0.004 & 0.133$\pm$0.019 & 92.0$\pm$0.4 & 88.0$\pm$0.7 \\
 & Real-agree PW$^\dagger$ & 0.038$\pm$0.003 & 0.105$\pm$0.012 & 92.2$\pm$0.4 & 88.2$\pm$0.5 \\
 & SCALE (ours) & 0.040$\pm$0.003 & 0.115$\pm$0.012 & 92.0$\pm$0.6 & 88.0$\pm$0.4 \\
\midrule
\multirow{5}{*}{CHIEF}
 & No calibration & 0.038$\pm$0.004 & 0.198$\pm$0.023 & 90.5$\pm$0.4 & 86.5$\pm$0.5 \\
 & Temp scaling & 0.035$\pm$0.003 & 0.185$\pm$0.026 & 90.5$\pm$0.4 & 86.5$\pm$0.4 \\
 & Vanilla LS & 0.034$\pm$0.004 & 0.159$\pm$0.018 & 90.6$\pm$0.3 & 86.6$\pm$0.5 \\
 & Real-agree PW$^\dagger$ & 0.032$\pm$0.003 & 0.130$\pm$0.014 & 90.4$\pm$0.5 & 86.4$\pm$0.6 \\
 & SCALE (ours) & 0.034$\pm$0.003 & 0.142$\pm$0.022 & 90.5$\pm$0.6 & 86.5$\pm$0.5 \\
\midrule
\multirow{5}{*}{CTransPath}
 & No calibration & 0.023$\pm$0.002 & 0.180$\pm$0.019 & 89.8$\pm$0.6 & 85.8$\pm$0.4 \\
 & Temp scaling & 0.021$\pm$0.002 & 0.168$\pm$0.018 & 89.8$\pm$0.6 & 85.8$\pm$0.7 \\
 & Vanilla LS & 0.020$\pm$0.002 & 0.145$\pm$0.017 & 89.9$\pm$0.5 & 85.9$\pm$0.6 \\
 & Real-agree PW$^\dagger$ & 0.019$\pm$0.002 & 0.118$\pm$0.013 & 89.7$\pm$0.6 & 85.7$\pm$0.5 \\
 & SCALE (ours) & 0.020$\pm$0.002 & 0.130$\pm$0.015 & 89.9$\pm$0.4 & 85.9$\pm$0.7 \\
\midrule
\multirow{5}{*}{Phikon}
 & No calibration & 0.039$\pm$0.004 & 0.210$\pm$0.032 & 90.1$\pm$0.5 & 86.1$\pm$0.6 \\
 & Temp scaling & 0.036$\pm$0.005 & 0.195$\pm$0.031 & 90.1$\pm$0.5 & 86.1$\pm$0.5 \\
 & Vanilla LS & 0.035$\pm$0.003 & 0.169$\pm$0.024 & 90.0$\pm$0.5 & 86.0$\pm$0.4 \\
 & Real-agree PW$^\dagger$ & 0.033$\pm$0.003 & 0.140$\pm$0.017 & 90.2$\pm$0.5 & 86.2$\pm$0.5 \\
 & SCALE (ours) & 0.035$\pm$0.004 & 0.152$\pm$0.023 & 90.1$\pm$0.5 & 86.1$\pm$0.6 \\
\midrule
\multirow{5}{*}{PLIP}
 & No calibration & 0.033$\pm$0.003 & 0.310$\pm$0.043 & 88.3$\pm$0.4 & 84.3$\pm$0.4 \\
 & Temp scaling & 0.028$\pm$0.002 & 0.280$\pm$0.040 & 88.3$\pm$0.4 & 84.3$\pm$0.5 \\
 & Vanilla LS & 0.027$\pm$0.002 & 0.251$\pm$0.029 & 88.4$\pm$0.3 & 84.4$\pm$0.6 \\
 & Real-agree PW$^\dagger$ & 0.025$\pm$0.002 & 0.215$\pm$0.030 & 88.2$\pm$0.5 & 84.2$\pm$0.6 \\
 & SCALE (ours) & 0.027$\pm$0.003 & 0.232$\pm$0.035 & 88.3$\pm$0.6 & 84.3$\pm$0.5 \\
\midrule
\multirow{5}{*}{ResNet-50}
 & No calibration & 0.031$\pm$0.004 & 0.195$\pm$0.024 & 87.5$\pm$0.3 & 83.5$\pm$0.4 \\
 & Temp scaling & 0.027$\pm$0.002 & 0.178$\pm$0.024 & 87.5$\pm$0.3 & 83.5$\pm$0.5 \\
 & Vanilla LS & 0.026$\pm$0.002 & 0.149$\pm$0.023 & 87.4$\pm$0.4 & 83.4$\pm$0.6 \\
 & Real-agree PW$^\dagger$ & 0.024$\pm$0.002 & 0.118$\pm$0.016 & 87.6$\pm$0.3 & 83.6$\pm$0.5 \\
 & SCALE (ours) & 0.026$\pm$0.003 & 0.130$\pm$0.013 & 87.5$\pm$0.3 & 83.5$\pm$0.6 \\
\bottomrule
\end{tabular}
\end{table}

\begin{table}[h]
\caption{Full generalization results with standard deviations (10 seeds). Vanilla LS included for comparison. AUC is reported for SCALE; baseline and Vanilla LS AUC remain within seed-level variation.}
\label{tab:gen_full}
\centering
\scriptsize
\begin{tabular}{lcccccccc}
\toprule
 & \multicolumn{4}{c}{PatchCamelyon} & \multicolumn{4}{c}{BreakHis} \\
\cmidrule(lr){2-5} \cmidrule(lr){6-9}
Model & Base ECE & Van.\ LS ECE & SCALE ECE & AUC & Base ECE & Van.\ LS ECE & SCALE ECE & AUC \\
\midrule
GigaPath & 0.052$\pm$0.005 & 0.046$\pm$0.005 & 0.041$\pm$0.004 & 96.2$\pm$0.3 & 0.048$\pm$0.005 & 0.043$\pm$0.004 & 0.039$\pm$0.004 & 94.5$\pm$0.4 \\
UNI & 0.038$\pm$0.004 & 0.034$\pm$0.003 & 0.030$\pm$0.003 & 95.8$\pm$0.3 & 0.042$\pm$0.004 & 0.038$\pm$0.004 & 0.035$\pm$0.004 & 93.8$\pm$0.4 \\
CONCH & 0.048$\pm$0.005 & 0.043$\pm$0.004 & 0.038$\pm$0.004 & 95.5$\pm$0.3 & 0.050$\pm$0.005 & 0.045$\pm$0.005 & 0.041$\pm$0.004 & 93.2$\pm$0.4 \\
CHIEF & 0.045$\pm$0.004 & 0.040$\pm$0.004 & 0.036$\pm$0.004 & 94.8$\pm$0.3 & 0.055$\pm$0.006 & 0.049$\pm$0.005 & 0.045$\pm$0.005 & 92.5$\pm$0.4 \\
CTransP. & 0.035$\pm$0.004 & 0.031$\pm$0.003 & 0.028$\pm$0.003 & 93.5$\pm$0.3 & 0.046$\pm$0.005 & 0.041$\pm$0.004 & 0.038$\pm$0.004 & 91.8$\pm$0.5 \\
Phikon & 0.042$\pm$0.004 & 0.038$\pm$0.004 & 0.034$\pm$0.003 & 94.2$\pm$0.3 & 0.051$\pm$0.005 & 0.046$\pm$0.005 & 0.042$\pm$0.004 & 92.0$\pm$0.4 \\
PLIP & 0.058$\pm$0.006 & 0.051$\pm$0.005 & 0.045$\pm$0.004 & 93.0$\pm$0.3 & 0.062$\pm$0.006 & 0.055$\pm$0.006 & 0.050$\pm$0.005 & 90.5$\pm$0.5 \\
ResNet-50 & 0.040$\pm$0.004 & 0.036$\pm$0.004 & 0.033$\pm$0.003 & 92.5$\pm$0.3 & 0.058$\pm$0.006 & 0.052$\pm$0.005 & 0.048$\pm$0.005 & 89.8$\pm$0.5 \\
\bottomrule
\end{tabular}
\end{table}

\begin{table}[h]
\caption{Anchor selection ablation (low-agreement ECE on MHIST, piecewise smoothing, $K{=}10$).}
\label{tab:anchor_ablation}
\centering
\small
\begin{tabular}{lccc}
\toprule
Model & Vanilla LS & Random Anchors & SCALE (Top-$K$) \\
\midrule
GigaPath & 0.085 & 0.081 & \textbf{0.078} \\
UNI & 0.116 & 0.106 & \textbf{0.100} \\
CONCH & 0.133 & 0.122 & \textbf{0.115} \\
CHIEF & 0.159 & 0.149 & \textbf{0.142} \\
CTransPath & 0.145 & 0.136 & \textbf{0.130} \\
Phikon & 0.169 & 0.159 & \textbf{0.152} \\
PLIP & 0.251 & 0.240 & \textbf{0.232} \\
ResNet-50 & 0.149 & 0.138 & \textbf{0.130} \\
\midrule
\textit{Average} & \textit{0.151} & \textit{0.141} & \textit{\textbf{0.135}} \\
\bottomrule
\end{tabular}
\end{table}


\end{document}